%% file: main.tex
\documentclass{article}

\usepackage{PRIMEarxiv}

\input{ide_preamble}

\usepackage[utf8]{inputenc} 
\usepackage[T1]{fontenc}    
\usepackage{hyperref}       
\usepackage{url}            
\usepackage{booktabs}       
\usepackage{amsfonts}       
\usepackage{nicefrac}       
\usepackage{microtype}      
\usepackage{lipsum}
\usepackage{fancyhdr}       
\usepackage{graphicx}       

\title{Improving Transformers using Faithful Positional Encoding
}

\author{Tsuyoshi~Id\'{e} \\
    \texttt{ tide@us.ibm.com} \\
    IBM T.~J.~Watson Research Center\\
    USA\\
    \And
    Jokin Labaien\\
    \texttt{jlabaien@ikerlan.es} \\
    Ikerlan Research Center\\
    Spain
    \And
    Pin-Yu Chen \\
    \texttt{pin-yu.chen@ibm.com}\\
    IBM T.~J.~Watson Research Center\\
    USA \\
    }

\begin{document}
\maketitle

\begin{abstract}
We propose a new positional encoding method for a neural network architecture called the Transformer. Unlike the standard sinusoidal positional encoding, our approach is based on solid mathematical grounds and has a guarantee of not losing information about the positional order of the input sequence. We show that the new encoding approach systematically improves the prediction performance in the time-series classification task.
\end{abstract}


\section{Introduction}

The ``Transformer''~\cite{vaswani2017attention} is one of the most successful neural network architectures for sequential data to date. Since the first proposal in 2016, it has been used in numerous tasks involving sequential data, and significant performance gains have been reported.

The Transformer is an algorithm to convert a sequence of raw data representation vectors $\{\bmx^{(1)}, \ldots, \bmx^{(S)}\}$ into another more informative data representation $\{\bmz^{(1)}, \ldots, \bmz^{(S)}\}$, where $S$ is the length (i.e., the number of items) of the sequence. We denote the dimensionality of the representation vectors by $\bmx^{(s)}\in \mathbb{R}^D$ and $\bmz^{(s)}\in \mathbb{R}^d$ with $s \in \{1,\ldots,S\}$. For the data representation to be informative, $\bmz^{(s)}$s have to reflect (1) the order and (2) the inter-dependency of the items. These two requirements are addressed by the following two steps in the Transformer algorithm~\cite{phuong2022formal}:

\begin{itemize}

\item Positional encoding, which blends in the information of the order to the raw representation.

\item Self-attention filter, which encourages dependent items to have similar representation vectors.

\end{itemize}

Typically, positional encoding (PE) is done by simply adding an extra vector $\bme^{(s)}$ to $\bmx^{(s)}$. For this extra vector $\bme^{(s)}$, many of the Transformer architectures proposed to date follow the original paper~\cite{vaswani2017attention}, which uses a vector of sinusoidal elements without giving clear justification.

Since the choice of $\bme^{(s)}$ seems somewhat arbitrary, a few attempts have been made to improve the Transformer with different forms of $\bme^{(s)}$. For instance,~\cite{shaw2018self} proposed the concept called relative PE, which modifies the value vector in the self-attention filter by incorporating the relative position of the items. \cite{raffel2020exploring} further simplified the relative PE algorithm by directly adding an extra term to the $(i,j)$ element of the self-attention matrix in the form of $b_{i-j}$. \cite{ke2021rethinking} proposes a variant of the relative PE that takes account of the asymmetry between the query and key matrices. Along with those handcrafted adjustments to PE, another popular direction is to introduce learnable parameters to the PE vector $\bme^{(s)}$. Examples include~\cite{lin2021pre}, which uses a sinusoidal vector with learnable frequencies. Also,~\cite{shukla2020multi} combines a sinusoidal function with a linear function to define extra dimensions to be concatenated with the raw representation vector.

Although those works report promising results in their settings, their justification boils down to the empirical evaluation on specific datasets chosen. The use of multiple datasets should enhance objectivity to some extent, but finite datasets can never cover the entire world. We need a principled way of analyzing and even deriving positional encoding.

To address this issue, we introduce a new notion of faithfulness and derive a new PE in a principled way as a solution that meets the faithfulness property. Our strategy is clear. We think of PE as a mapping from a localized position function (a.k.a. one-hot vector) to a smoother differentiable function. If there exists an inverse mapping, the PE holds the full information of the position function and hence can be said to be faithful. As suggested by the sinusoidal form that has been used since the original PE, we leverage the theory of discrete Fourier transform to derive a new PE method called the DFT encoding that meets the faithfulness requirement. We confirm that the newly developed faithful PE improves the classification task performance. To the best of our knowledge, this is the first work that derived an optimal PE based on the notion of faithfulness.

\section{Preliminary}

This section provides a summary of the original Transformer algorithm and the basics of discrete Fourier transform (DFT), which plays a critical role in deriving the proposed PE approach.

\subsection{Problem setting}

As introduced in the previous section, the goal of the Transformer algorithm~\cite{vaswani2017attention} is to ``transform'' an input vector sequence into another sequence of enriched representation vectors. We assume that the input sequence has $S$ \textit{items}, and each of the items has a raw data representation vector. An item can be a word or a time-series segment, and the representation vector can be the one-hot vector of a word or a vector of observed values in a time-series segment. 

We denote the raw data representation by $\bmx^{(s)} \in \mathbb{R}^D$ and the enriched representation by $\bmz^{(s)} \in \mathbb{R}^d$ for $s=1,\ldots,S$. The transformer can be viewed also as a transformation from an input data matrix of raw data representation vectors $\sfX \triangleq [\bmx^{(1)},\ldots, \bmx^{(S)}] \in \mathbb{R}^{D\times S}$ into another data matrix of new data representation vectors $\sfZ \triangleq [\bmz^{(1)},\ldots, \bmz^{(S)}]  \in \mathbb{R}^{d \times S}$. The motivation behind this transformation is to get a richer representation that captures not only the information of the word itself but also the dependency between the words. In other words, $\bmz^{(s)}$ carries information not only about the $s$-th word but also the other words that are semantically related. 

In what follows, $d$ is assumed to be even but extension to the case where $d$ is odd is straightforward. All the vectors are column vectors and are denoted in bold italic. Matrices are denoted in sans serif.

\subsection{The Transformer algorithm}

The Transformer algorithm consists of two steps: Positional encoding (PE) and self-attention filtering.

\subsubsection{Positional encoding}

The goal of PE is to reflect the information of the position of the items (i.e., words in machine translation) in the input sequence. In the original formulation~\cite{vaswani2017attention}, this was done by simply adding an extra vector to the raw representation vector of an input word: 
\begin{align} \label{eq:positional_x+e}
    {\bmx}^{(s)} \leftarrow \bmx^{(s)}  + \bme^{(s)},
\end{align}
where $\bmx^{(s)}$ is the raw representation vector of the $s$-th item and $\bme^{(s)}$ is its corresponding positional encoding vector. \cite{vaswani2017attention} used
\begin{align}\label{eq:positional_encoding_vaswani}
    \bme^{(s)} = \tilde{\bme}^{(s)} \triangleq
    & \left(\sin (w_0 s),  \cos (w_0 s), \sin (w_2 s), \cos (w_2 s), \ldots, 
    \sin (w_{d-2} s), \cos (w_{d-2} s)   \right)^\top,
\end{align}
where
\begin{align}\label{eq:VanilaPE_wk}
     w_k \triangleq \rho^{-\frac{k}{d}}, \quad k=0, 2, 4, \ldots, d-2.
\end{align}
In \cite{vaswani2017attention}, $\rho$ is a hard-coded parameter. They chose $\rho=10000$ and $d =512$ possibly through empirical results out of a few candidate values. This specific sinusoidal form seems to have been based on an intuition to capture different length scales of word-word dependency.

\subsubsection{Self-attention filtering}

With Eq.~\eqref{eq:positional_x+e}, we now have an updated data matrix $\sfX$. The self-attention filtering further update $\sfX$ to get $\sfZ$. There are three steps in the algorithm. The first step is to create three different feature vectors from $\sfX$:
\begin{align}
    \sfQ \triangleq \sfW_Q\sfX, \ \sfK\triangleq\sfW_K\sfX,\  \sfV \triangleq \sfW_V\sfX,
\end{align}
which are called the query, key, and value respectively. The matrices $\sfW_Q, \sfW_K, \sfW_V$ are $d \times D$ and are to be learned from data. Denote the column vectors as $\sfQ = [\bmq^{(1)},\ldots, \bmq^{(S)}]$, $\sfK = [\bmk^{(1)},\ldots, \bmk^{(S)}]$, and $\sfV = [\bmv^{(1)},\ldots, \bmv^{(S)}]$. Note that $\sfQ, \sfK, \sfV$ as well as $\sfX$ have been defined as column-based data matrices.

The second step is to compute the self-attention matrix $\sfA$. From the query and key matrix, the self-attention matrix $\sfA \in \mathbb{R}^{S \times S}$ is defined as
\begin{align}
    \sfA \triangleq \mathrm{softmax}\left(\frac{1}{\sqrt{d}}\sfQ^\top \sfK\right),
\end{align}
where the softmax function applies to each row. To write down the $(i,j)$ element explicitly, we have
\begin{align}
    A_{i,j} = \frac{\exp (B_{i,j}) }{ \sum_{m=1}^S \exp(B_{i,m}) }, \quad B_{i,j} \triangleq \frac{1}{\sqrt{d}}(\bmq^{(i)})^\top \bmk^{(j)}.
\end{align}
This definition implies that $\sfA$ is essentially a cosine-similarity matrix between the items $i$ and $j$. Hence,if, e.g.,~$A_{1,3}$ and $A_{1,6}$ dominate the first row, the 3rd and 6th items can be viewed as item 1's neighbors.

Finally, the third step is to take in the influence of neighbors in each row:
\begin{align}
    \bmz^{(s)} = \sum_{k=1}^S A_{s,k} \bmv^{(k)} \quad \mbox{or} \quad 
    \sfZ = \sfV \sfA^\top
\end{align}
Obviously, through this step, the representation vectors are encouraged to have similar values to their neighbors. 

Typically, self-attention filtering is multiplexed by using multiple ``heads.''~\footnote{The term ``head'' seems to be originated from the Turing machine, where a head means a tape head to read magnetic tapes.} Let $H$ be the number of heads. In this approach, $H$ different sets of the parameter matrices are used $\{\sfW_Q^{[h]}, \sfW_K^{[h]}, \sfW_V^{[h]}\mid h =1,\ldots, H \}$ with different initializations. As a result, $H$ different self-attention matrices $\{\sfA^{[h]}\}$ and value vectors $\{\bmv^{(s)[h]} = \sfW_V^{[h]}\bmx^{(s)}\}$ are obtained. For each, the representation vectors are computed as 
\begin{align}
    \bmz^{(s)[1]} =  \sum_{k=1}^S A_{s,k}^{[1]} \bmv^{(k)[1]},\quad \ldots, 
    \quad \bmz^{(s)[H]} =  \sum_{k=1}^S A_{s,k}^{[H]}\bmv^{(k)[H]} \quad \in \mathbb{R}^{d}.
\end{align}
The final representation is created simply by concatenating these $H$ vectors as
\begin{align}
    \bmz^{(1)} = \begin{pmatrix}
    \bmz^{(1)[1]}\\ \vdots\\ \bmz^{(1)[H]}
    \end{pmatrix}, \quad \ldots, \quad 
    \bmz^{(S)} = \begin{pmatrix}
    \bmz^{(S)[1]}\\ \vdots\\ \bmz^{(S)[H]}
    \end{pmatrix}\quad \in \mathbb{R}^{dH}.
\end{align}

\subsection{Discrete Fourier transform (DFT)}

The Transformer algorithm is concerned with $d$-dimensional representation vectors representing the items in the sequence of length $S$. Finding a $d$-dimensional representation vector can be viewed as assigning a function defined on a one-dimensional lattice with $d$ lattice points. This is a useful change of view since we have a rich set of mathematical tools to study the expressiveness of functions. In this subsection, we recapitulate the basics of DFT, which plays a central role in the proposed DFT PE.

It is well-known that \textit{any} function $f(t)$ defined on the $d$-dimensional lattice $t = 0,1,\ldots, d-1$ can be expanded with the discrete Fourier bases as  
\begin{align}\label{eq:realDFT_general}
    f(t) = \frac{1}{\sqrt{d}}[a_0 + b_0 \cos (\pi t)]+  \sqrt{\frac{2}{d}}\sum_{k=1}^K\left( a_k\cos (\omega_k t)
    +b_k \sin (\omega_k t) \right),
\end{align}
where we have assumed that $d$ is \textit{even} and
\begin{align}
    K \triangleq \frac{d}{2} -1, \quad \omega_k \triangleq \frac{2\pi k}{d}. 
\end{align}
The expansion~\eqref{eq:realDFT_general} defines a mapping from the function $f(t)$ to the set of coefficients $\{(a_l,b_l)\mid l=0,\ldots,K\}$. This mapping is called the discrete Fourier transform (DFT). Notice that the total number of terms is $2+2K=d$. For notational simplicity, let us define
\begin{align*}
    \varphi_l(t) \triangleq
    \begin{cases}
    \frac{1}{\sqrt{d}} & l=0\\
    \sqrt{\frac{2}{d}}\cos (\omega_l t) & l=1,\ldots,K\\
    \sqrt{\frac{2}{d}}\sin (\omega_{l-K} t) & l=K+1,\ldots,2K\\
    \frac{1}{\sqrt{d}}\cos (\pi t) &l=d-1
    \end{cases}.
\end{align*}
It is straightforward to verify that these basis functions satisfy the orthogonality condition

\begin{align}
\sum_{t=0}^{d-1} \varphi_l(t)\varphi_m(t)=\delta_{l,m},
\end{align}

where $\delta_{l,m}$ is Kronecker's delta. The orthogonality of the Fourier bases allows us to easily find the coefficients as

\begin{align}\label{eq:DFT-varphi}
f(t) = \sum_{l=0}^{d-1} c_l \varphi_l(t), \quad \quad \mbox{where}\quad
c_l = \sum_{t=0}^{d-1}f(t)\varphi_l(t).
\end{align}

In terms of $a_l$s and $b_l$s, we have

\begin{align}\label{eq:DFT-varphi_ab}
a_0 = \sum_{t=0}^{d-1}f(t)\varphi_0(t), \quad
b_0 = \sum_{t=0}^{d-1}f(t)\varphi_{d-1}(t), \quad
a_l = \sum_{t=0}^{d-1}f(t)\varphi_l(t), \quad
b_l = \sum_{t=0}^{d-1}f(t)\varphi_{l+K}(t)
\end{align}

for $l=1,\ldots,K$. For more details of the relationship between real and complex DFTs, see, e.g.~\cite{ersoy1985real}.

One of the most remarkable properties of DFT is that DFT is \textit{invertible}. In the DFT expansion~\eqref{eq:DFT-varphi}, the function values $\bmf \triangleq (f(0),\ldots,f(d-1))^\top$ can be fully recovered by $\bmc \triangleq (c_0,\ldots,c_{d-1})^\top$, given the DFT bases. In other words, $\bmc$ and $\bmf$ have exactly the same information; one can think of $\{c_l\}$ as another representation of $f(t)$.

\section{Fourier Analysis of Positional Encoding}

In this section, we discuss how Fourier analysis is used to evaluate the goodness of positional encoding. 

\subsection{Original PE has a strong low-pass property}

Positional encoding is the task of finding a $d$-dimensional representation vector of an item in the input sequence of length $S$. In the vector view, the position of the $s$-th item in the input sequence is most straightfowardly represented by an $S$-dimensional ``one-hot'' vector, whose $s$-th entry is 1 and otherwise 0. Unfortunately, this is not an appropriate representation because the dimensionality is 
fixed to be $S$, and it does not have continuity over the elements at all, which makes numerical optimization challenging in stochastic gradient descent. 

The original PE in Eq.~\eqref{eq:positional_encoding_vaswani} is designed to address these limitations. Then, the question is how we can evaluate the goodness of its specific functional form. One approach suggested by the sinusoidal form is to use DFT. As before, consider DFT on the 1-dimensional (1D) lattice with $d$ lattice points. As shown in Eq.~\eqref{eq:realDFT_general}, any function can be represented as a linear combination of sinusoidal functions with frequencies $\{\omega_k\}$. It is interesting to see how the frequencies $\{w_k\}$ in Eq.~\eqref{eq:VanilaPE_wk} are distributed in the Fourier space.

\begin{figure}[bth]
\centering    
\includegraphics[trim={0cm 0cm 0cm 0cm},clip,width=10cm]{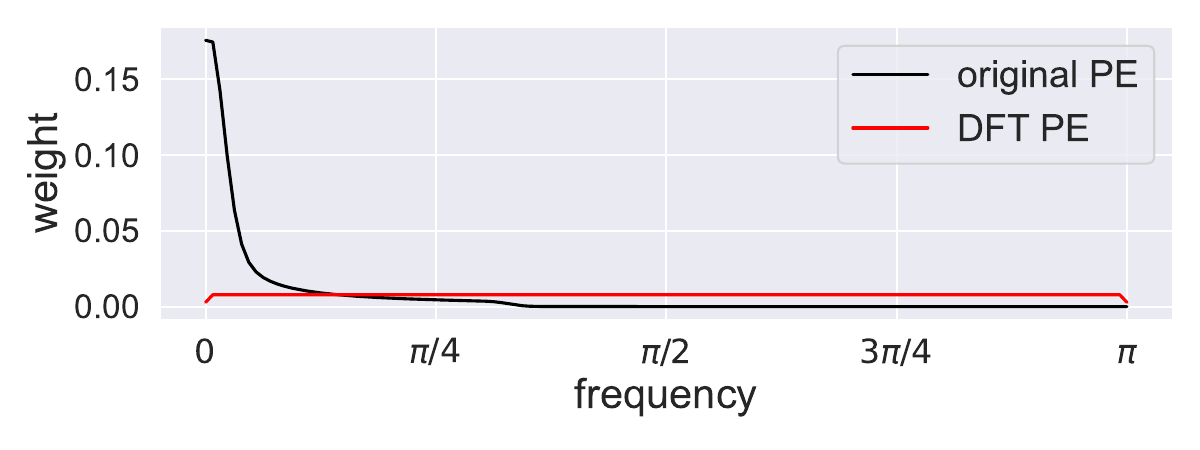}
    \caption{The distribution of the frequency $w_k = \rho^{-\frac{k}{d}}$ in Eq.~\eqref{eq:VanilaPE_wk} over the Fourier bases (black line), where $\rho=10000,d=256$. Notice the contrast to that of the DFT encoding, which gives a uniform distribution (red line; See Section~\ref{sec:DFT_encoding}).}
    \label{fig:frequencyRatio}
    \vspace{-0.0cm}
\vspace{-0.0cm}
\end{figure}

The black line in Fig.~\ref{fig:frequencyRatio} shows the distribution of the frequencies in Eq.~\eqref{eq:VanilaPE_wk}. The distribution is estimated using kernel density estimation with the Gaussian kernel~\cite{Bishop}. Specifically, at each $\omega_k$, the weight is given by
\begin{align}
g_k &= \sum_{l \in \{0,2,\ldots,d-2 \}}\frac{1}{R}\exp\left(-\frac{1}{2 \sigma^2}(\omega_k - w_l)^2\right),
\end{align}
where $R$ is a normalization constant that ensures $\sum_k g_k=1$. We chose the bandwidth $\sigma=4\times \frac{2\pi}{d}$ with $d=256$. Due to the power function $\rho^{-\frac{k}{d}}$, the distribution is extremely skewed towards zero.

This fact can also be easily understood by conducting a simple analysis as follows. The first and second smallest frequencies are $0$ and $\frac{2\pi}{d}$, respectively. We can count the number of $w_k$s that fall between them. Solving the equation
\begin{align}
\frac{2\pi}{d} = \rho^{-\frac{l}{d}}
\end{align}
and assuming $\rho=10000$, we have $l = \frac{d}{4}\log_{10}\frac{d}{2\pi} \approx 103$ for $d=256$ and $l\approx 245$ for $d=512$. Hence, almost half of the entries go into this lowest bin.

This simple analysis, along with Fig.~\ref{fig:frequencyRatio}, demonstrates that the original PE strongly suppresses mid and high frequencies. As a result, it tends to ignore mid- and short-range differences in location. This can be problematic in applications where short-range dependencies matter, such as time-series classification for physical sensor data.

\subsection{Original PE lacks faithfulness}

Another interesting question is what kind of function the skewed distribution $g_k$ represents. To answer this question, we perform an experiment described in Algorithm~\ref{algo:reference_reconstruction}, which is designed to understand what kind of distortion it may introduce to an assumed reference function. In our PE context, the reference function should be the position function (a.k.a.~one-hot vector) since the original PE~\eqref{eq:positional_encoding_vaswani} was proposed to be a representation of the item at the location $s$.

\begin{algorithm}[tb]
\caption{Reference function reconstruction}
\label{algo:reference_reconstruction}
\begin{algorithmic}[1] 
\REQUIRE Reference function $f(t)$, DFT component weights $\{g_k\}$.
\STATE Find DFT coefficients of $f$ as $\{(a_k,b_k)\mid k=0,\ldots,K\}$.
\STATE $a_0 \leftarrow a_0 g_0$ and $b_0 \leftarrow b_0 g_{K+1}$
\FORALL{$k=1,\ldots,K$}
\STATE $a_k \leftarrow a_k g_k$ and $b_k \leftarrow b_k g_k$
\ENDFOR
\STATE Inverse-DFT from the modified coefficients.
\end{algorithmic}
\end{algorithm}

\begin{figure}[bth]
\centering    
\includegraphics[trim={0cm 0cm 0cm 0cm},clip,width=10cm]{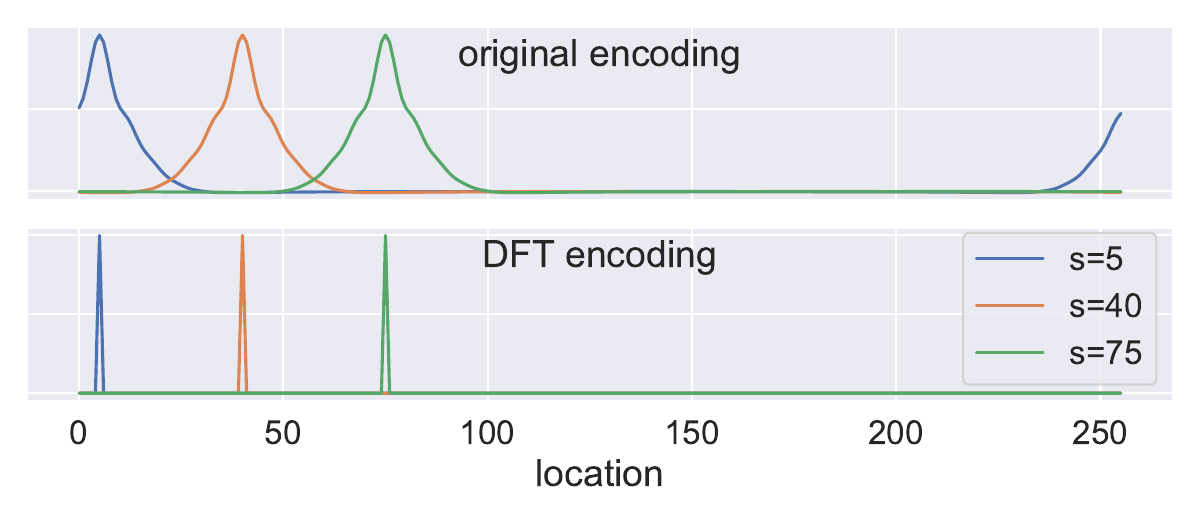}
    \caption{Reconstruction by Algorithm~\ref{algo:reference_reconstruction} for the location function $f(t)=\delta_{t,5},\ \delta_{t,40}, \ \delta_{t,75}$. Perfect reconstruction corresponds to single-peaked spikes at $t=5, 40, 75$, respectively. The broad distributions given by the original PE (top) demonstrate a significant loss of information in the original PE. See Section~\ref{sec:DFT_encoding} for DFT encoding (bottom). }
    \label{fig:reconstruction_comparison}
    \vspace{-0.0cm}
\vspace{-0.0cm}
\end{figure}

Figure~\ref{fig:reconstruction_comparison} shows the result of reconstruction. We normalized the modified DFT coefficients so the original $\ell_2$ norm is kept unchanged. All the parameters used are the same as those in Fig.~\ref{fig:frequencyRatio}, i.e.,~$d=256,S=80,h=10000, \sigma=4\times \frac{2\pi}{d}$. As expected from the low-pass property, the reconstruction by the original PE failed to reproduce the delta functions. The broad distributions imply that the original PE is not sensitive to the difference in the location up to about 30. As the total sequence length is $S=80$, we conclude that the original PE tends to put an extremely strong emphasis on global long-range dependencies within the sequence. 

If the PE is supposed to represent the position function, PE should faithfully reproduce the original position function. Here, we formally define the notion of the faithfulness of PE:
\begin{definition}[Faithfulness of PE]
A positional encoding is said to be faithful if it is injective (i.e., one-to-one) to the position function.
\end{definition}

For PE, the requirement of faithfulness seems natural. Interestingly, as long as the sinusoidal bases are assumed, this requirement almost automatically leads to a specific PE algorithm that we call the DFT encoding, which is the topic of the next section.

\section{DFT Positional Encoding}\label{sec:DFT_encoding}

For the requirement of faithfulness, one straightforward approach is to leverage DFT for positional encoding. The idea is to use the DFT representation as a smooth surrogate for the one-hot function $f_s(t) \triangleq \delta_{s,t}$. Let $a_0^{(s)},a_1^{(s)},\ldots, b_1^{(s)}, \ldots, b_K^{(s)}, b_0^{(s)}$ be the DFT coefficients of $f_s(t)$. It is straightforward to compute the coefficients against the real Fourier bases:
\begin{align}\label{eq:DFT_def_a0}
    a_0^{(s)} &= \sum_{t=0}^{d-1} \delta_{s,t}\varphi_0(t)=\sum_{t=0}^{d-1} \delta_{s,t}\frac{1}{\sqrt{d}}
    =\frac{1}{\sqrt{d}}
    \\ \label{eq:DFT_def_a1}
    a_1^{(s)} &= \sum_{t=0}^{d-1} \delta_{s,t}\varphi_1(t)=\sum_{t=0}^{d-1} \delta_{s,t}\sqrt{\frac{2}{d}}\cos(\omega_1 t) =\sqrt{\frac{2}{d}}\cos(\omega_1 s)
    \\ \nonumber
     &\vdots
     \\ \label{eq:DFT_def_bK}
    b_K^{(s)}&= \sum_{t=0}^{d-1} \delta_{s,t}\varphi_K(t)=\sum_{t=0}^{d-1} \delta_{s,t}\sqrt{\frac{2}{d}}\sin(\omega_K t) =\sqrt{\frac{2}{d}}\sin(\omega_K s)
    \\ \label{eq:DFT_def_b0}
    b_0^{(s)}&=  \sum_{t=0}^{d-1} \delta_{s,t}\varphi_{d-1}(t)=\sum_{t=0}^{d-1} \delta_{s,t}\sqrt{\frac{1}{d}}\cos(\pi t) =\sqrt{\frac{1}{d}}\cos(\pi s)
\end{align}
which provides the vector representation of the one-hot function as
\begin{align}\label{eq:DFT_def_summary}
\bme^{(s)} \triangleq (a_0^{(s)},a_1^{(s)},\ldots, b_1^{(s)}, \ldots, b_K^{(s)}, b_0^{(s)})^\top.
\end{align}
We call this the \textbf{DFT encoding}.


Because of the general properties of DFT, the following claim is almost evident: 
\begin{theorem}
The DFT encoding is faithful. 
\end{theorem}
(Proof) This follows from the existence of the inverse transformation in DFT. We can also prove this directly. By using Eq.~\eqref{eq:DFT_def_summary}, the r.h.s.~of the DFT expansion in Eq.~\eqref{eq:realDFT_general} will be
\begin{align*}
    \mbox{r.h.s.}&= \frac{1}{d}[1 + \cos (\pi t) \cos (\pi s)]
    + \frac{2}{d}\sum_{k=1}^K\left[\cos(\omega_k s)\cos(\omega_k t) + \sin(\omega_k s)\sin(\omega_k t)\right]
    \\
    &=  \frac{1}{d}[1 + (-1)^{s-t}]
    + \frac{2}{d}\sum_{k=1}^K \cos(\omega_k (s-t))
\end{align*}
It is obvious that the r.h.s.~is 1 if $s=t$. Now, assume $s\neq t$. Using $\cos x =\frac{1}{2}(\rme^{\rmi x} + \rme^{-\rmi x})$ and the sum rule of geometric series, we have
\begin{align*}
     \mbox{r.h.s.}&= \frac{1}{d}[1 + (-1)^{s-t}] + \frac{1}{d}\left[
     \frac{c_{s,t} - c_{s,t}^{K+1}}{1-c_{s,t}} + \frac{c_{s,t}^{-1} - c_{s,t}^{-K-1}}{1 - c_{s,t}^{-1}}
     \right]
     = \frac{1}{d}[1 + (-1)^{s-t}] + \frac{1}{d}
     \frac{c_{s,t} - c_{s,t}^{K+1}  -1 + c_{s,t}^{-K}}{1-c_{s,t}} 
\end{align*}
where we have defined $c_{s,t} \triangleq \exp\left(\rmi \frac{2\pi(s-t)}{d}\right)$. Noting $c_{s,t}^{-2K-1} = c_{s,t}$ and $c_{s,t}^{K+1} = (-1)^{s-t}$, we have
\begin{align*}
     \mbox{r.h.s.}&= \frac{1}{d}[1 + (-1)^{s-t}] -     \frac{1}{d} [1 + (-1)^{s-t}] = 0.
\end{align*}
Putting all together, the inverse DFT of the DFT encoding gives $\delta_{s,t}$, which is the location function.\qedsymbol

In Fig.~\ref{fig:frequencyRatio}, we have shown the distribution of the DFT encoding in the Fourier domain. From Eqs.~\eqref{eq:DFT_def_a0}-\eqref{eq:DFT_def_b0}, we see that the distribution is given by
\begin{align}
    g_k^{\mathrm{DFT}} = \frac{1}{d}(\delta_{k,0} + \delta_{k,d-1})
    +\frac{2}{d}\sum_{l=1}^K\delta_{k,l}
    = \frac{1}{d}(\delta_{k,0} + \delta_{k,d-1})
    +\frac{2}{d},
\end{align}
where $\delta_{k,0}$ etc.~are Kronecker's delta function. This distribution is flat except for the terminal points at $k=0,d-1$. Hence, unlike the original PE, the proposed DFT PE does not have any bias on the choice of the Fourier components. 

With this flat distribution, we also did the reconstruction experiment. The result is shown in the bottom panel in Fig.~\ref{fig:reconstruction_comparison}. We see that the location functions are perfectly reconstructed with the DFT PE.

\section{Experimental evaluation}

We tested DFT PE in a time-series classification task. The input is a multivariate time-series and the output is a binary label representing whether the input time-series is anomalous or not. We used three benchmark datasets: Elevator, SMD, and MSL. The results are summarized in Table~\ref{tab:empirical_evaluation}, where F1 is the harmonic average between precision and recall.

\begin{table}[thb]
\centering
\caption{Comparison between the original PE and DFT PE in a time-series classification task on three benchmark data sets: Elevator, SMD, and MSL. DFT PE consistently gives better performance. } \begin{tabular}{c|c c c | c c c }
    \hline \hline
        &  & original PE & &  & DFT PE & \\
     & precision & recall & F1 &  precision & recall & F1 \\
     \hline
    Elevator & 0.977 & 0.917 & 0.946 & 0.979 & 0.955 & \textbf{0.967}\\ 
    SMD & 1 & 0.943 & 0.970 & 1 & 0.961 & \textbf{0.980}\\
    MSL & 0.822 & 0.855 & 0.838 & 0.894 & 0.821 & \textbf{0.856}\\
    \hline
    \end{tabular}
    \label{tab:empirical_evaluation}
\end{table}

For more details of the experimental evaluation, see our companion paper~\cite{labaien2023diagnostic}.

\section{Conclusion}

We have proposed a new positional encoding method for the Transformer. The main contribution of this paper is to establish DFT as the theoretical basis for positional encoding. Using DFT, we discussed in what sense the original PE can be suboptimal. Based on the new notion of faithfulness in PE, we derived a new approach named DFT PE. In the time-series classification task, we confirmed that DFT PE systematically improved the performance, possibly because of the importance of short- and mid-range dependencies, which have been ignored by the original PE.


\bibliographystyle{apalike}
\bibliography{ide_et_al}

\end{document}

%% file: ide_preamble.tex
\usepackage{algorithm}
\usepackage{algorithmic}
\usepackage{wrapfig}
\usepackage{graphicx}
\usepackage{mathtools} 
\usepackage{amssymb}
\usepackage{amsmath}
\usepackage{amsthm} 
\usepackage{color}
\usepackage{bm}

\newtheorem{theorem}{Theorem}

\newtheorem{definition}{Definition}
\usepackage{tcolorbox}

\newcommand{\bmc}{\bm{c}}

\newcommand{\bme}{\bm{e}}
\newcommand{\bmf}{\bm{f}}

\newcommand{\bmk}{\bm{k}}

\newcommand{\bmq}{\bm{q}}

\newcommand{\bmv}{\bm{v}}

\newcommand{\bmx}{\bm{x}}

\newcommand{\bmz}{\bm{z}}

\newcommand{\rme}{\mathrm{e}}

\newcommand{\rmi}{\mathrm{i}}

\newcommand{\sfA}{\mathsf{A}}

\newcommand{\sfK}{\mathsf{K}}

\newcommand{\sfQ}{\mathsf{Q}}

\newcommand{\sfV}{\mathsf{V}}
\newcommand{\sfW}{\mathsf{W}}
\newcommand{\sfX}{\mathsf{X}}

\newcommand{\sfZ}{\mathsf{Z}}